\documentclass[10pt,twocolumn,letterpaper]{article}

\usepackage{cvpr}      
\definecolor{cvprblue}{rgb}{0.21,0.49,0.74}
\usepackage[pagebackref,breaklinks,colorlinks,allcolors=cvprblue]{hyperref}
\usepackage{tabularx}
\usepackage{xcolor}
\usepackage{colortbl}
\usepackage{xcolor}  
\usepackage{tcolorbox}  %
\tcbuselibrary{listings} 
\newtcolorbox{promptbox}[1][]{
  colback=gray!10,    
  colframe=gray!50,   
  boxrule=0.5pt,      
  arc=2pt,            
  left=8pt, right=8pt, top=6pt, bottom=6pt,
  title=#1
}
\usepackage{amsmath}
\usepackage{multirow}
\def\confName{CVPR}
\def\confYear{2026}

\title{Black-box Membership Inference Attacks on the \\ Pre-training Data of Image-generation Models}

\author{Tao Qi$^{1}$\thanks{Equal contribution.}
 \quad Huili Wang$^{2}$\footnotemark[1]  \thanks{Corresponding Author. (Email: wwwanghuili@gmail.com)}\quad Yuanhong Huang$^{1}$\footnotemark[1] \quad Wendan Wang$^{1}$ \\
Lianchao Zhao$^{1}$ \quad Jinrui Wang$^{1}$ \quad Zichen Qin$^{1}$ \quad Shangguang Wang$^{1}$ \quad Yongfeng Huang$^{2}$ \\
$^{1}$ Beijing University of Posts and Telecommunications \quad $^{2}$ Tsinghua University}

\begin{document}
\maketitle
\begin{abstract}

The rapid advancement of diffusion-based image generation models has raised serious concerns regarding potential copyright and privacy infringements involving human-created data.
Membership inference attacks (MIAs) have emerged as a promising tool for identifying unauthorized data usage during model training.
Existing methods typically assess the ability of model to denoise perturbed suspect images as an indicator of membership status.
However, the discriminative power of such features is highly dependent on the degree of model memorization and deteriorates significantly when applied to less exposed data (e.g., pre-training data).
Although several methods attempt to enhance detection by leveraging internal model features, these features are generally inaccessible in mainstream closed-source image generation platforms, limiting their practicality.
In this paper, we demonstrate that analyzing how a black-box diffusion model denoises a target image and corresponding perturbed textual instructions can reveal more distinctive membership cues. 
Based on this insight, we propose a black-box membership inference attack framework (named SD-MIA) that leverages a cross-modal data perturbation mechanism to detect pre-training data in diffusion models.
We conduct extensive experiments on both a public benchmark dataset and a newly constructed dataset, each comprising pre-training membership and non-membership samples with identical distributions. 
Experimental results demonstrate that SD-MIA achieves superior performance compared to existing baselines, including those with the unfair advantage of accessing internal model features. 

\end{abstract}    
\section{Introduction}
\label{sec:intro}


Although diffusion-based generative models have achieved remarkable success in image synthesis~\cite{dubinski2024towards,esser2024scaling}, they have also raised substantial concerns regarding the unauthorized use of copyrighted data~\cite{sag2023copyright, zhang2024copyright, carlini2021extracting,karamolegkou2023copyright}.
To prevent such data misuse, membership inference attack methods~\cite{shokri2017membership,hu2022membership} have been developed and widely adopted to detect whether specific data were used in the training of diffusion models.
The mainstream paradigm of existing MIAs typically involves perturbing a suspected image with random noise and then measuring the model denoising capability as an indicator of membership~\cite{pangblack,zhai2024membership}.
Besides, to enable controlled evaluation of MIA performance, prior work commonly fine-tunes open-source diffusion models on randomly partitioned datasets and assesses attacks within these synthetic settings~\cite{matsumoto2023membership,duan2023diffusion,kongefficient,fu2025unlocking}.
Since the fine-tuned model typically exhibits pronounced memorization of the newly introduced fine-tuning data, these methods usually report strong detection accuracy.
However, such results do not reflect real-world deployments, where the vast majority of training data is consumed during large-scale pre-training and remains substantially less exposed to the attacker.
Recent studies further demonstrate that image perturbation–based detection signals, while effective for fine-tuning data, deteriorate sharply when applied to pre-training samples~\cite{dubinski2024towards,liang2024real}.
Consequently, existing methodologies become largely impractical for detecting unauthorized data usage in realistic diffusion model scenarios.


Several recent studies have attempted to improve membership-inference performance on pre-training data. These approaches continue to rely on perturbation-induced visual signals, but further incorporate fine-grained internal features extracted from the denoising trajectory to better characterize behavioral discrepancies between members and non-members~\cite{kongefficient,li2024unveiling}.
For instance, \citet{kongefficient} leverage intermediate noise predictions along the diffusion process to compare reconstructed and expected noise patterns for membership detection.
However, practical image generation systems~\cite{betker2023improving,saharia2022photorealistic} typically expose only final outputs and do not provide access to intermediate model computation states, making such gray-box techniques infeasible in real-world settings.
This gap highlights the need for a principled black-box MIA method that can reliably operate on pre-training data in image generation models.


We begin by revisiting the well-documented failure of image-space perturbation methods~\cite{ho2020denoising,rombach2022high}, which are effective for fine-tuning data attacks but do not generalize to the pre-training setting. The core issue arises from two structural properties of modern diffusion pipelines: the VAE encoder contracts away fine-grained image variations, and the stochastic denoising trajectory quickly extinguishes any residual perturbation signal~\cite{kingma2013auto,ho2020denoising,rombach2022high}, making members and non-members display almost indistinguishable reconstruction behavior under visual space perturbation.
However, these observations motivate a shift toward the textual modality in our work. Since text embeddings remain unnoised throughout generation, and diffusion models usually internalize locally overfitted textual–visual mappings for pre-training samples, small textual perturbations for member textual descriptions tend to remain within a collapsed representation region and thus yield stable generations. By contrast, textual perturbations applied to non-member data are more likely to exit this region, producing noticeably divergent outputs and thereby offering informative signals.

Building on this structural asymmetry, we introduce a black-box membership inference framework (named SD-MIA) for identifying pre-training data of diffusion-based image generators.
First, we design a multi-view textual perturbation strategy that induces controlled embedding shifts under black-box constraints while preserving instruction coherence. 
Second, to convert these perturbation-induced behaviors into a discriminative membership signal, we develop a cross-modal relevance–based estimator of the unobservable generation probability, together with a maximum-relevance pooling mechanism that mitigates diffusion stochasticity and amplifies the behavioral divergence between members and non-members.
We conduct comprehensive experiments on two benchmarks to evaluate the performance of our MIA framework in detecting pre-training data within diffusion models.
The first benchmark, LAION-mi~\cite{dubinski2024towards}, contains pre-training samples from Stable Diffusion v1.2–v1.5 and non-membership samples with an aligned distribution.
To further assess generalization to more recent architectures, we extend this benchmark to Stable Diffusion v3.5 and a set of mainstream closed-source image generation models, including DALL·E, OpenAI, and Gemini, enabling evaluation across multiple model generations.
Experimental results on both benchmarks consistently demonstrate that SD-MIA significantly outperforms state-of-the-art MIA methods.
Details of codes and datasets are released at \url{https://github.com/wanghl21/SD-MIA}.


\section{Related Work}
\label{sec:related}

Membership inference attacks (MIAs) on image-generation models have recently attracted increasing attention, aiming to determine whether a given sample was involved in model training. 
Depending on the level of access by adversaries, existing MIAs can be broadly categorized into {white-box}, {gray-box}, and {black-box} paradigms (Table~\ref{Tab:relatedwork} in Appendix). 

White-box and gray-box methods typically exploit internal model signals, such as gradients, latent noise predictions, or reconstruction losses along the denoising trajectory to identify memorized samples~\cite{pang2023white,matsumoto2023membership,duan2023diffusion,fu2025unlocking}. For instance, \citet{fu2025unlocking} leveraged generative priors to show that member samples are reconstructed more faithfully from degraded inputs. 
While these methods have achieved promising results for overfitted fine-tuning datasets, multiple studies~\cite{dubinski2024towards,liang2024real} consistently show that their effectiveness sharply declines in large-scale pre-training settings, where memorization is weaker and internal activations are inaccessible. 
A few recent gray-box attempts~\cite{kongefficient,fu2025unlocking,zhai2024membership} have explored MIAs against pre-training data by leveraging latent generative priors. However, these methods still rely on semi-transparent access to internal features, making them impractical for commercial diffusion systems. 
In contrast, our work introduces SD-MIA, a fully black-box framework for pre-training membership inference. 

\begin{figure*}[ht]
  \centering
   \includegraphics[width=0.98\linewidth]{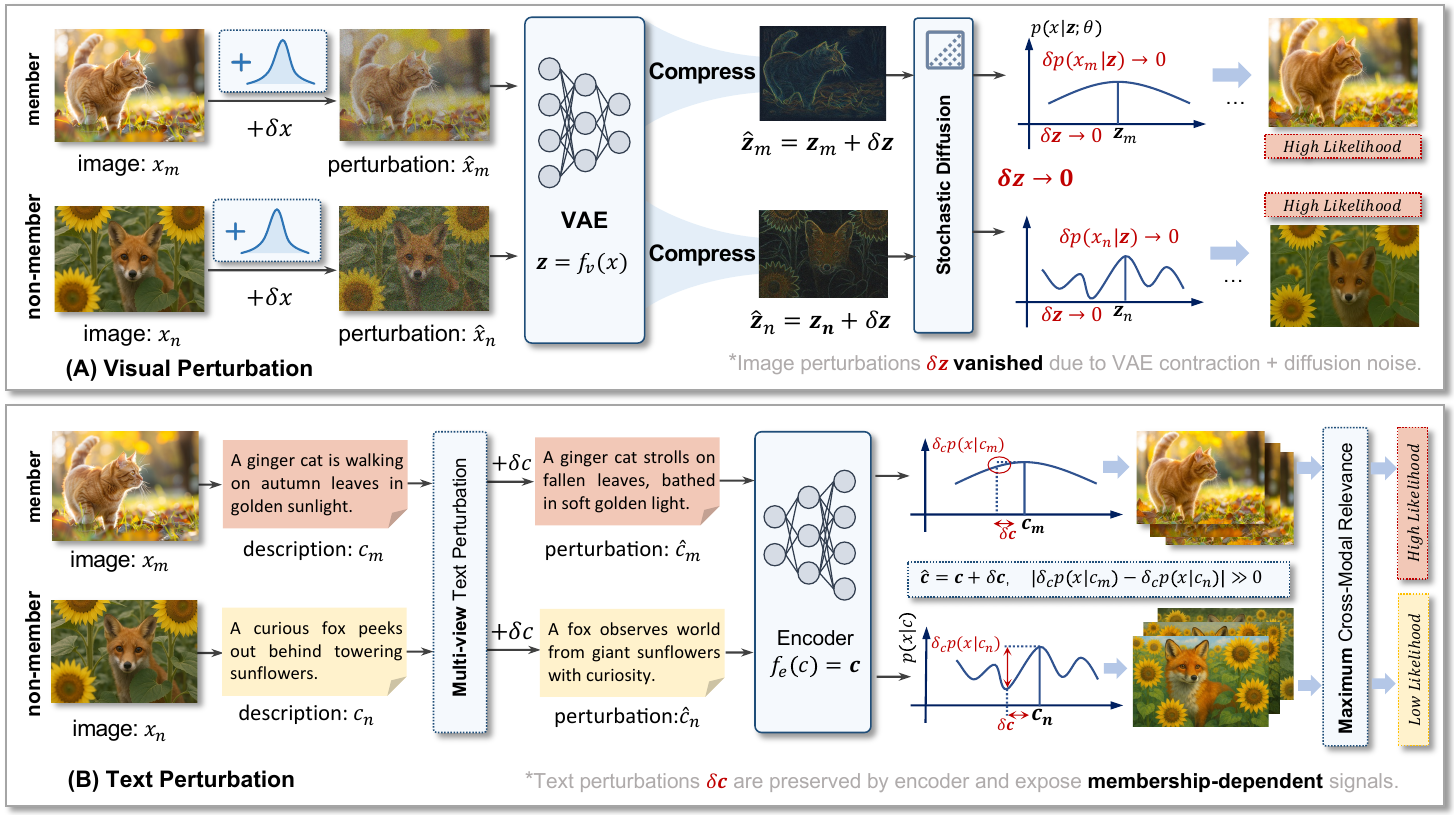}
   \caption{
Methodology insights and framework.
(A) Visual perturbation struggles to reveal the generation probability curvature gap between member and non-member, resulting in insignificant detection signals.
(B) SD-MIA addresses this via a multi-view text perturbation mechanism, enabling precise measurement of curvature shifts and reliable detection of pre-training data in diffusion models.}
   \label{fig:method}
\end{figure*}

Existing MIA evaluation protocols also face significant limitations in assessing pre-training membership. 
Most prior studies validate their attacks on fine-tuning-based setups, where diffusion models are retrained on partitioned datasets to simulate member and non-member samples~\cite{matsumoto2023membership,duan2023diffusion}. 
Although convenient, such settings may induce artificial overfitting and exaggerate memorization effects, leading to overestimated attack performance that fails to reflect real-world pre-training dynamics. 
Some recent works~\cite{kongefficient, zhai2024membership,fu2025unlocking} attempt to evaluate MIAs directly on pre-training data. However, their non-member selection procedures often introduce distributional biases, as non-members are drawn from unrelated datasets, thereby compromising fairness and interpretability. 
For example, \citet{zhai2024membership} selected non-members from the MS-COCO dataset to contrast against LAION-based pre-training members, resulting in domain-level discrepancies in object composition and textual description density that inadvertently simplify the classification task.
To ensure unbiased evaluation, we adopt the LAION-mi benchmark~\cite{dubinski2024towards}, which aligns member and non-member distributions under the same data domain. 
Furthermore, we extend this benchmark to include both open-source and closed-source models, to audit the large-scale pre-training in realistic diffusion model deployments.

\section{Methodology}
\label{sec:method}

\subsection{Adversarial Assumptions}
Following the standard black-box threat model for diffusion MIAs~\cite{li2025towards,pangblack}, we assume that the adversary can only interact with $\mathcal{M}_\theta$ via query-based text-to-image generation.
The adversary does not have access to the target model parameters, training data, optimization procedure, 
or any intermediate latent representations. 
Instead, the attacker can 
issue a (possibly adaptive) sequence of textual prompts $ \{c_1, c_2, \ldots, c_K\}$ 
and obtain the corresponding generated images $\hat{x}_i \sim \mathcal{M}_\theta(\cdot \mid c_i)$. 
This setup reflects the practical constraints of commercial text-to-image APIs, 
where only input–output access is permitted.

\subsection{Motivation of SD-MIA}

\begin{figure}[h]
    \centering
    \includegraphics[width=1.0\linewidth]{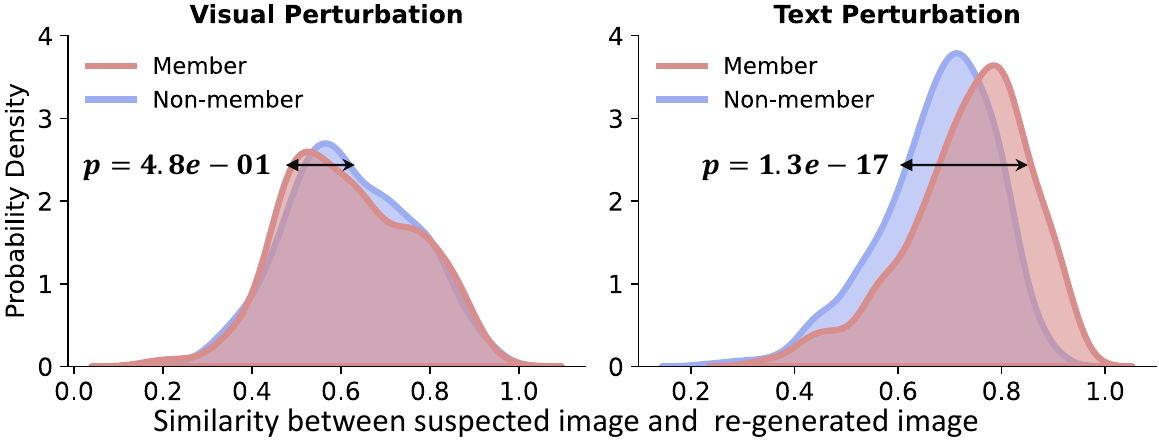}
    \caption{Distributional comparison between visual-perturbation method and text-perturbation black-box method SD-MIA.}
    \label{fig:score_distribution}
\end{figure}

\textbf{Adaption of existing methodology.}
To determine whether a suspected image appeared in the pre-training corpus of a black-box image generative model, an intuitive strategy is to extend membership-inference techniques originally developed for fine-tuning data~\cite{fu2025unlocking,duan2023diffusion}. However, our empirical evaluation (Figure~\ref{fig:score_distribution} and Table~\ref{Tab:main}) shows that such extensions fail to generalize effectively. Their central limitation lies in dependence on image-space perturbation and denoising-response analysis.
In modern diffusion-based generators~\cite{ramesh2022hierarchical,zhang2023adding,saharia2022photorealistic}, input images are first encoded into variational latents that aggressively compress fine-grained structure and smooth away moderate perturbations. These latents are then propagated through a stochastic diffusion trajectory, where injected noise rapidly overwhelms any remaining residual variation~\cite{ho2020denoising,rombach2022high}. As a result, perturbations applied in the image domain rarely survive the generative process, causing member and non-member samples to exhibit similarly stable reconstructions (Fig.~\ref{fig:method} A).
Formally, let $c$ denote the textual description and $x$ the target image. Consider the change in the (unobservable) generation probability $p(x| c;\theta^*)$ when perturbing $x$ by $\delta x$. Let $\mathbf z = f_v(x)$ be the VAE latent encoding of $x$, $\mathbf c = f_e(c)$ the embedding of the text description, and $\theta^*$ for parameters. The induced variation $\delta_x p$ satisfies:
\begin{equation}
\begin{aligned}
    \delta_x p
    &\approx \bigl| \nabla_{\mathbf z} p(\mathbf z,\mathbf c;\theta^*) \cdot \delta \mathbf z \bigr| \\
    &\lesssim \bigl\| \nabla_{\mathbf z} p(\mathbf z,\mathbf c;\theta^*) \bigr\|_2 \cdot \bigl\|\delta \mathbf z \bigr\|_2 .
\end{aligned}
\label{eq:delta_p}
\end{equation}
Since VAE encoders exhibit locally contractive behavior~\cite{kingma2013auto,rolinek2019variational}, we further have:
\begin{equation}
    \bigl\|\delta\mathbf z\bigr\|_2
    = \bigl\| f_v(x+\delta x) - f_v(x) \bigr\|_2
    \lesssim \bigl\| J_{f_v}(x) \bigr\|_2 \, \bigl\| \delta x \bigr\|_2 ,
\label{eq:delta_z}
\end{equation}
where $J_{f_v}(x)$ denotes the Jacobian of the encoder at $x$. Since $\|J_{f_v}(x)\|_2 \ll 1$ for typical VAEs, the latent perturbation $\|\delta\mathbf z\|_2$ quickly vanishes even when $|\delta x|_2$ is non-negligible. The stochastic diffusion dynamics further attenuate this already minuscule latent variation, driving $\delta_x p$ toward zero regardless of potential differences in $\|\nabla_{\mathbf z} p\|$ between member and non-member cases:
\begin{equation}
    |\delta_x p(x_m) -\delta_{{x}}p(x_n)|\approx \xi\cdot \delta x \rightarrow 0 , \quad |J_{f_v}(x)|<\xi .
\end{equation}
Moreover, generation models rarely exhibit strong memorization of individual pre-training images, which further limits the magnitude of image perturbation-induced variations. Crucially, in black-box scenarios, the quantity $\delta_x p$ is itself unobservable, and any attempt to approximate it by analyzing generated samples introduces additional stochastic variability. This noise further masks the already subtle discrepancies between member and non-member inputs, rendering image perturbation–based inference unreliable.

\textbf{Our insight.} In contrast, we argue that a {cross-modal perturbation-reconstruction} mechanism offers a more reliable pathway for identifying pre-training membership. Instead of perturbing the image itself, we perturb the textual instructions that condition the generator and evaluate the reconstruction consistency of the target image (Figure~\ref{fig:method} B). Crucially, text and images follow fundamentally different computational pathways in diffusion generators: textual instructions are mapped into representations that remain unnoised throughout the entire generation process, persisting as conditioning signals that guide the denoising trajectory.
During training, the model usually internalizes a locally overfitted mapping from the representation of a specific description, and its immediate representation neighborhood, into the visual features of a particular training sample. This induces a {representation-region collapse}, where a neighborhood of semantically proximate text variants funnels into a single visual mode. For member images, small textual perturbations remain within this collapsed region, producing outputs that remain close to the memorized appearance. For non-member data, no such collapsed structure exists, and perturbations shift the conditioning signal to disparate regions of the representation space, producing outputs with noticeably higher variability. 
Here, we provide a more formal analysis.  When perturbing the description $c$ by $\delta c$, the induced variation in the generation probability satisfies:
\begin{equation}
\begin{aligned}
    \delta_c p
    &\approx \bigl| \nabla_{\mathbf c} p(\mathbf z,\mathbf c;\theta^*) \cdot \delta \mathbf c \bigr| \\
    &\lesssim \bigl\| \nabla_{\mathbf c} p(\mathbf z,\mathbf c;\theta^*) \bigr\|_2 \cdot \bigl\|\delta \mathbf c \bigr\|_2 .
\end{aligned}
\label{eq:delta_c}
\end{equation}
Under representation region collapse, member pairs $(x_m,c_m)$ satisfy
$\bigl\| \nabla_{\mathbf c} p(\mathbf z_m,\mathbf c_m;\theta^*) \bigr\|_2 \approx 0,$ while this condition does not hold for non-member pairs $(x_n,c_n)$. Since textual perturbations induce observable shifts $\delta\mathbf c$ in the embedding space, we obtain $\delta_c p(x_m) \ll \delta_c p(x_n)$
for identical perturbation magnitudes $\|\delta\mathbf c\|_2$, providing a theoretical basis for using cross-modal perturbation as a discriminative signal for pre-training membership:
\begin{equation}
|\delta_cp(x_m)-\delta_{{c}}p(x_n)|\approx \bigl\| \nabla_{\mathbf c} p(\mathbf z_n,\mathbf c_n;\theta^*) \cdot \delta{\textbf{c}_n} \bigr\|_2 \gg 0.
\end{equation}
Furthermore, we conduct an empirical evaluation on stable-diffusion-v1.5 to support the analysis (Fig.~\ref{fig:score_distribution}).
For each suspected sample, we apply both visual and textual perturbations, prompt the model to reconstruct the image, and then compute the similarity between the original and regenerated outputs.
The results indicate that visual perturbations yield negligible distributional differences between member and non-member samples, whereas textual perturbations introduce pronounced and consistently separable shifts, highlighting their superior efficacy for probing membership signals.
Finally, we note that, to the best of our knowledge, a cross-modal perturbation-reconstruction mechanism has not been systematically conceptualized or instantiated in prior MIA studies for diffusion models.

\subsection{Indirect Textual Representation Perturbation}

Building on the preceding analysis, a core component of SD-MIA is the controlled perturbation of textual representations.
To exploit the representation-region collapse associated with member samples, the perturbations must satisfy two simultaneous requirements. 
First, the altered representations should remain within the collapsed neighborhood relevant to the suspected member.
Second, the perturbation should induce a measurable embedding displacement $\lVert\delta\mathbf c\rVert$ that allows us to probe whether reconstruction probability persists under controlled variation.
Since the black-box setting prevents direct manipulation of internal text embeddings, we construct a multi-view textual perturbation strategy that uses natural-language transformations as an indirect but structured mechanism for inducing embedding shifts. 
To ensure that these transformations remain coherent and preserve cross-modal consistency, a large language model (e.g., GPT-5~\cite{singh2025openaigpt5card}) is employed to generate linguistically diverse yet semantically regulated re-writings.

Specifically, we design three complementary perturbation views that span a graded spectrum of embedding displacements.
The first one is \emph{token-view perturbation.}
This view performs lexical and syntactic rewritings while preserving the semantic content and the essential descriptive intent of the original prompt. 
These minimally invasive changes induce fine-grained embedding displacements, enabling us to examine whether the representation remains within the collapsed region associated with member samples.
The second one is \emph{style-view perturbation.}
This view alters stylistic attributes such as register, density of description, narrative framing, or degree of elaboration, while keeping the semantic content intact. 
These perturbations create moderate representation shifts that test the extent to which the collapsed region exhibits stability against stylistic drift in the textual description.
The third one is \emph{semantic-view perturbation.}
This view introduces controlled modifications to semantic attributes (e.g., altering objects within images) while retaining the stylistic profile of the original text. 
These yield the strongest representation shifts and are designed to explore the boundary at which the perturbed representation exits the collapsed neighborhood and begins to produce divergent reconstructions, particularly for non-member samples where no collapsed region exists.
To prevent perturbations from drifting excessively far from the original description, we impose the similarity constraint on each perturbed textual description $\hat{c}$:
$
\mathrm{sim}\!\left( f_e(c), f_e(\hat{c}) \right) \ge \tau ,
$
where $\tau$ controls the upper bound on semantic deviation.

Using this mechanism, we obtain a set of perturbed textual descriptions
$
\{\hat{c}^{t}_i,\hat{c}^{s}_i,\hat{c}^{c}_i\}_{i=1}^N,
$
where $\hat{c}^{t}_i$, $\hat{c}^{s}_i$, and $\hat{c}^{c}_i$ denote the token-, style-, and semantic-view perturbations, respectively, and $N$ is the total number of perturbations.
By evaluating reconstruction behavior across this perturbation spectrum, SD-MIA enables a controlled and systematic examination of representation-region collapse in a black-box setting, thereby establishing a practical and theoretically grounded basis for inferring pre-training membership.

\begin{table*}[ht]
\centering
\caption{
Membership inference performance across different diffusion-based image generation models.
We consider balanced and imbalanced (1:10) proportion of member to non-member, and report the AUC and the true positive rate under a false positive rate of $\leq5\%$.
}

\label{Tab:main}
\resizebox{\textwidth}{!}{%
\begin{tabular}{c|cc|cc|cc|cc}
\toprule
\multirow{2}{*}{Methods} & \multicolumn{2}{c|}{Stable-diffusion-v1-2 (balanced)} & \multicolumn{2}{c|}{Stable-diffusion-v1-4 (balanced)} & \multicolumn{2}{c|}{Stable-diffusion-v1-5 (balanced)}  & \multicolumn{2}{c}{Stable-diffusion-v3-5 (balanced)} \\ 
                         & AUC          & TPR@5\% FPR         & AUC          & TPR@5\% FPR        & AUC          & TPR@5\% FPR         & AUC          & TPR@5\% FPR         \\ \midrule
Loss                     &  51.59 $\pm$ 1.94 & 4.73 $\pm$ 1.24 & 52.91 $\pm$ 1.38 & 6.53 $\pm$ 1.94 & 53.75 $\pm$ 3.85 & 6.53 $\pm$ 1.59 & 42.10 $\pm$ 1.30 & 1.20 $\pm$ 0.62 \\
SecMIA &
51.66 $\pm$ 1.89 & 5.87 $\pm$ 1.81 & 55.49 $\pm$ 1.39 & 5.13 $\pm$ 1.67 & 52.10 $\pm$ 2.77 & 5.53 $\pm$ 1.65 & 34.68 $\pm$ 2.18 & 2.07 $\pm$ 1.00 \\
PIA & 52.66 $\pm$ 2.43 & 7.20 $\pm$ 2.22 & 49.52 $\pm$ 2.87 & 6.47 $\pm$ 1.05 & 48.16 $\pm$ 2.49 & 7.00 $\pm$ 1.91 & 50.62 $\pm$ 1.53 & 2.40 $\pm$ 1.04  \\
CLiD & 49.26 $\pm$ 2.18 & 5.33 $\pm$ 1.92 & 53.71 $\pm$ 1.75 & 5.60 $\pm$ 3.23 & 51.88 $\pm$ 1.52 & 6.60 $\pm$ 1.72 & 58.15 $\pm$ 1.88 & 4.33 $\pm$ 1.01 \\
DRC & 54.66 $\pm$ 1.40 & 11.27 $\pm$ 2.24 & 55.83 $\pm$ 1.24 & 9.13 $\pm$ 3.36 & 54.61 $\pm$ 1.46 & 9.40 $\pm$ 1.51 & 60.44 $\pm$ 1.11 & 12.47 $\pm$ 1.42 \\ \midrule
REDIFFUSE  & 49.42 $\pm$ 1.29 & 4.53 $\pm$ 1.00 & 51.27 $\pm$ 3.45 & 5.67 $\pm$ 2.49 & 50.01 $\pm$ 2.05 & 4.93 $\pm$ 1.72 & 45.29 $\pm$ 1.94 & 3.00 $\pm$ 1.38 \\ 
Reconstruction  & 
59.66 $\pm$ 1.57 & 12.13 $\pm$ 1.44 & 60.99 $\pm$ 2.98 & 14.40 $\pm$ 4.06 & 60.30 $\pm$ 2.47 & 11.73 $\pm$ 3.70 & 46.74 $\pm$ 1.63 & 1.47 $\pm$ 0.45 \\ 
SD-MIA  & \cellcolor{gray!30} \textbf{66.28} $\pm$ 1.47 & \cellcolor{gray!30} \textbf{16.73} $\pm$ 2.98 & \cellcolor{gray!30} \textbf{66.23} $\pm$ 1.76 & \cellcolor{gray!30} \textbf{15.80} $\pm$ 3.98 & \cellcolor{gray!30} \textbf{65.92} $\pm$ 1.29 & \cellcolor{gray!30} \textbf{13.67} $\pm$ 3.37 & \cellcolor{gray!30} \textbf{66.93} $\pm$ 0.53 & \cellcolor{gray!30} \textbf{18.33} $\pm$ 1.19
\\ \midrule  \midrule
\multirow{2}{*}{Methods} & \multicolumn{2}{c|}{Stable-diffusion-v1-2 (imbalanced)} & \multicolumn{2}{c|}{Stable-diffusion-v1-4 (imbalanced)} & \multicolumn{2}{c|}{Stable-diffusion-v1-5 (imbalanced)}  & \multicolumn{2}{c}{Stable-diffusion-v3-5 (imbalanced)} \\ 
                         & AUC          & TPR@5\% FPR         & AUC          & TPR@5\% FPR         & AUC          & TPR@5\% FPR         & AUC          & TPR@5\% FPR        \\ \midrule
Loss & 52.98 $\pm$ 3.70 & 6.40 $\pm$ 2.33 & 52.88 $\pm$ 4.63 & 6.80 $\pm$ 2.99 & 53.05 $\pm$ 4.04 & 4.40 $\pm$ 1.50 & 41.04 $\pm$ 1.84 & 1.20 $\pm$ 1.60 \\
SecMIA & 51.78 $\pm$ 4.07 & 5.20 $\pm$ 2.04 & 55.75 $\pm$ 3.24 & 4.80 $\pm$ 2.04 & 54.03 $\pm$ 3.37 & 7.60 $\pm$ 1.96 & 33.26 $\pm$ 1.18 & 1.60 $\pm$ 1.50 \\
PIA & 51.14 $\pm$ 2.03 & 5.60 $\pm$ 1.50 & 45.84 $\pm$ 1.46 & 7.20 $\pm$ 0.98 & 48.94 $\pm$ 3.43 & 6.40 $\pm$ 2.65 & 49.06 $\pm$ 3.61 & 3.20 $\pm$ 2.04 \\
CLiD & 53.05 $\pm$ 2.06 & 5.20 $\pm$ 2.99 & 51.74 $\pm$ 1.39 & 5.20 $\pm$ 1.60 & 50.99 $\pm$ 2.04 & 5.20 $\pm$ 2.04 & 55.81 $\pm$ 1.69 & 4.80 $\pm$ 2.99 \\
DRC & 53.98 $\pm$ 2.25 & 10.40 $\pm$ 5.28 & 56.79 $\pm$ 2.90 & 12.80 $\pm$ 5.88 & 56.10 $\pm$ 3.92 & 8.40 $\pm$ 3.88 & 61.24 $\pm$ 7.91 & 18.40 $\pm$ 6.62 \\ \midrule
REDIFFUSE & 52.08 $\pm$ 1.70 & 6.40 $\pm$ 0.80 & 47.80 $\pm$ 3.42 & 3.60 $\pm$ 2.94 & 45.50 $\pm$ 4.40 & 4.00 $\pm$ 2.19 & 45.36 $\pm$ 3.10 & 3.48 $\pm$ 2.95 \\
Reconstruction  & 57.61 $\pm$ 5.26 & 17.20 $\pm$ 5.74 & 60.85 $\pm$ 3.56 & 16.00 $\pm$ 4.56 & 59.72 $\pm$ 3.50 & 10.80 $\pm$ 1.60 & 52.89 $\pm$ 2.96 & 4.40 $\pm$ 1.50 \\ 
SD-MIA  & \cellcolor{gray!30} \textbf{65.98} $\pm$ 3.46 & \cellcolor{gray!30} \textbf{13.62} $\pm$ 2.89 & \cellcolor{gray!30} \textbf{66.12} $\pm$ 2.78 & \cellcolor{gray!30} \textbf{17.02} $\pm$ 5.55 & \cellcolor{gray!30} \textbf{66.22} $\pm$ 5.79 & \cellcolor{gray!30} \textbf{14.04} $\pm$ 5.31 & \cellcolor{gray!30}  \textbf{65.04} $\pm$ 3.44 & \cellcolor{gray!30} \textbf{18.40} $\pm$ 6.12 \\ 
\bottomrule  
\end{tabular}
}
\end{table*}

\subsection{Maximum Cross-Modal Relevance Estimation}

In practical black-box settings, the generation probability 
$p(x | c; \theta^*)$ is inaccessible. 
To approximate this quantity, we introduce a \emph{maximum cross-modal relevance estimator} that amplifies observable probability curvature differences both within and beyond the collapsed representation region. 
This estimation is motivated by two observations.

First, although $p(x | \hat{c};\theta^*)$ cannot be queried, the relevance between the target image $x$ and the reconstruction $\hat{x}$ generated from a perturbed description $\hat{c}$ serves as a stable surrogate signal. 
Perturbations inside the collapsed region yield reconstructions closely aligned with $x$, whereas perturbations outside provoke semantic and structural deviations. 
Motivated by this property, we introduce a cross-modal relevance modeling approach to obtain fine-grained estimations of probability variation across heterogeneous semantic spaces (i.e., visual and textual domains). For each pair $(x,\hat{x})$, we first obtain textual descriptions $d_x$ and $d_{\hat{x}}$ by an image caption model, then embed both images and their summaries, and finally compute a joint relevance score:
\begin{equation}
    s(x,\hat{c}) 
    = \bigl( h_v(x) \oplus h_t(d_{x}) \bigr)
      \cdot 
      \bigl( h_v(\hat{x}) \oplus h_t(d_{\hat{x}}) \bigr),
\end{equation}
where $h_v(\cdot)$ and $h_t(\cdot)$ denote visual and textual encoders (e.g. CLIP), and $\oplus$ denotes the operation of feature concatenation. 
The resulting score $s(x,\hat{c})$ serves as a principled surrogate for the unobservable probability  $p(x|\hat{c};\theta^*)$.

Second, the stochastic nature of diffusion-based generation injects considerable variability into relevance-based probability estimation.
To counteract this, we leverage an additional structural asymmetry. 
For non-member data, the model lacks any memorized correspondence between the conditioning description and the target image, making faithful reconstruction of $x$ effectively unattainable, even after extensive resampling. 
By contrast, for member data, the model can reproduce the target image with non-negligible probability when queried repeatedly.
This divergence in reproducibility motivates a {maximum-relevance} mechanism that accentuates the separation between member and non-member signals by pooling the highest relevance values obtained across multiple stochastic reconstructions.

Concretely, for a suspected pair $(c,x)$, we perturb data $c$ under a given view (e.g., token level), reconstruct the image, evaluate its cross-modal relevance with the target image, and pool the top $K\%$ of relevance scores:
\begin{equation}
    s^t = \frac{1}{n} {\textstyle \sum_{j=1}^{n}} 
     s\bigl(x,\hat{c}^{t}_{R_j}\bigr), \quad
    n = \left\lfloor N \cdot K\% \right\rfloor,
\end{equation}
where $s^t$ is the maximum relevance score for the token-view perturbations, 
$\{\hat{c}^{t}_i\}_{i=1}^{N}$ is the perturbation set,
$R_j$ indexes the top-ranked relevance values, and $K\%$ is the pooling proportion.
Analogously, we compute maximum relevance scores for the style- and semantic-view perturbations, denoted $s^s$ and $s^c$, respectively.
These view-specific maxima are then integrated into a unified maximum cross-modal relevance estimator $s_f(x,\hat{c})$, which provides a stable surrogate for $p(x,\hat{c}| \theta^*)$. In parallel, a corresponding estimator $s_f(x,c)$ can be derived to approximate the unperturbed likelihood $p(x,{c}| \theta^*)$.
Finally, we compute $s_f = s_f(x,\hat{c}) - s_f(x, c)$ as an empirical approximation of $\delta_c p$.
This quantity reflects text perturbation-induced shifts in the generation probability and provides an effective estimate of the underlying probability curvature, thereby enabling reliable black-box pre-training membership inference.


\section{Experiment}
\label{sec:experiment}

\subsection{Experimental setup}

\textbf{Datasets and models.}
To ensure a comprehensive and standardized evaluation, we follow the established LAION-mi benchmark~\cite{dubinski2024towards}, which provides a realistic and fair setting for membership inference without requiring any fine-tuning or post-training. Under this protocol, we evaluate our attack on several widely used text-to-image diffusion models, including stable-diffusion series~\cite{Rombach_2022_CVPR}, targeting membership signals arising from their pretraining datasets. 
To emulate realistic settings, where model parameters and pretraining data are fully inaccessible, we further construct a new benchmark FlickrMIA-25 following the methodology of~\cite{dubinski2024towards}.  
Member samples are sourced from LAION-2B, while non-member samples are drawn from Flickr, restricted to images released after January 1st, 2025, to reflect temporal disjointness from the training corpora of current models.
This benchmark enables evaluation on both stable-diffusion-3.5-large~\cite{esser2024scaling}, as well as commercial black-box systems, including DALL·E~\cite{betker2023improving}, GPT-4o~\cite{islam2025gpt}, and Gemini~\cite{gemini2flash_modelcard_2025}, providing a realistic test of membership inference performance in practical, closed-source environments.
We additionally apply embedding-level distribution matching filters to reduce dataset bias.
The resulting data distribution is visualized in Figure~\ref{fig:data_distribution} (see Appendix~\ref{App:Unbiasedness} for details).



\begin{figure}[ht]
\centering\includegraphics[width=1\linewidth]{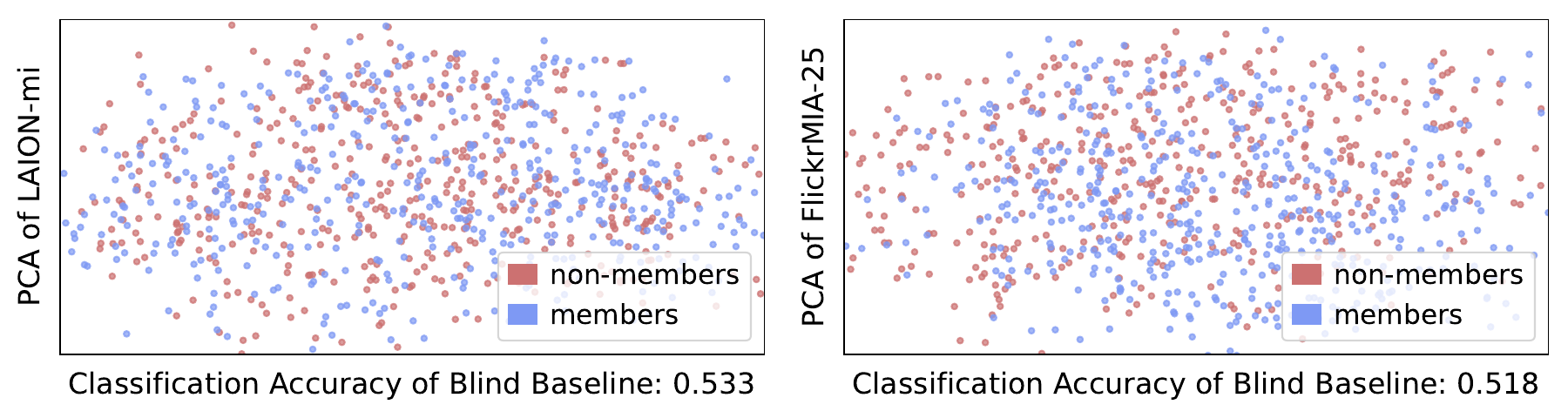}
    \caption{Aligned distributions and near-chance classifier accuracy demonstrate the absence of benchmark bias.}
    \label{fig:data_distribution}
\end{figure}

\noindent\textbf{Baseline and implementation settings.}
We benchmark SD-MIA against state-of-the-art MIA approaches for diffusion models. The baselines span different adversarial access levels, including white-box, gray-box, and black-box methods. 
Details of seven baseline methods are in the Appendix~\ref{App:baseline}.
We use CLIP ViT-L/14~\cite{radford2021learning} to extract both image and text embeddings. When a description is unavailable, we generate surrogate one using Blip2-opt-6.7b~\cite{li2023blip}, and all perturbations are produced with GPT-5~\cite{singh2025openaigpt5card} (see Appendix~\ref{App:prompt} for the full prompt set). The perturbation thresholds are fixed as follows: $\tau_t = 0.9$ for token-view perturbations, $\tau_s = 0.8$ for style-view perturbations, and $\tau_c = 0.6$ for semantic-view perturbations.
For each perturbed description, the diffusion model is queried 10 times to obtain multiple reconstructions.
To reduce stochastic variation, all experiments are repeated five times with different random seeds. We report the mean and standard deviation for standard evaluation metrics, including the area under the ROC curve (AUC) and TPR@5\% FPR. In addition to instance-level evaluation, we also assess set-level membership inference, where AUC measures how well the true training set can be distinguished from non-training sets. This setting reflects more realistic threat models in which the adversary aims to determine whether a specific dataset has contributed to the model pretraining process. 

\begin{figure*}
    \centering
    \includegraphics[width=0.95\linewidth]{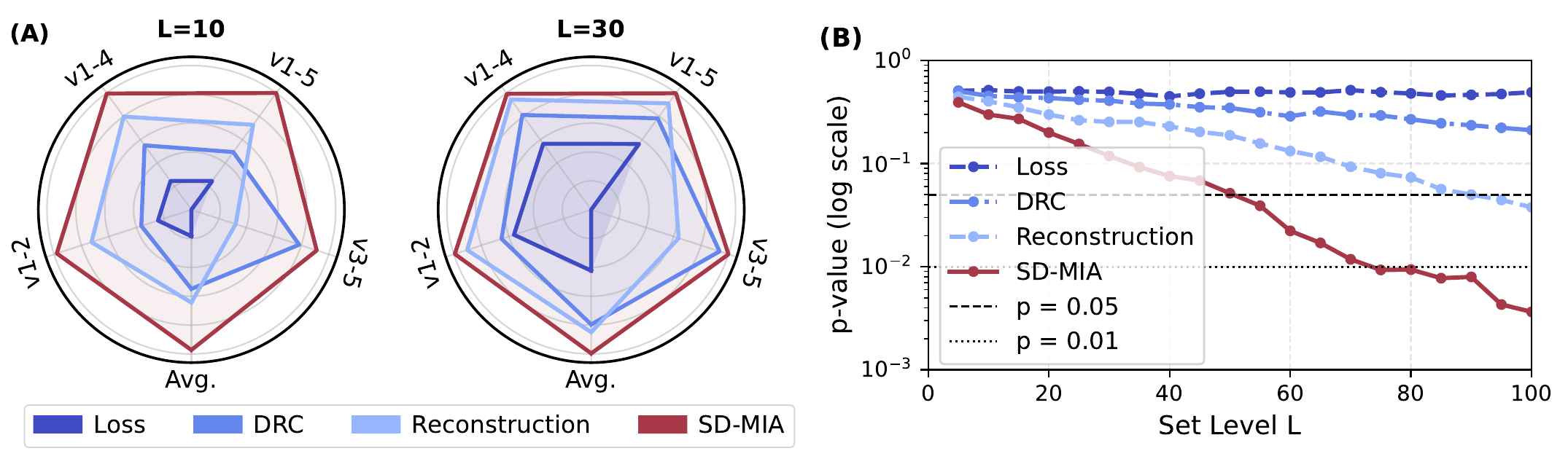}
    \caption{Set-level MIA performance: (A) AUC results at different set sizes $L$; (B) corresponding p-values for sd v1–4 across $L$.}
    \label{fig:set-level}
\end{figure*}

\begin{figure*}
    \centering
    \includegraphics[width=0.95\linewidth]{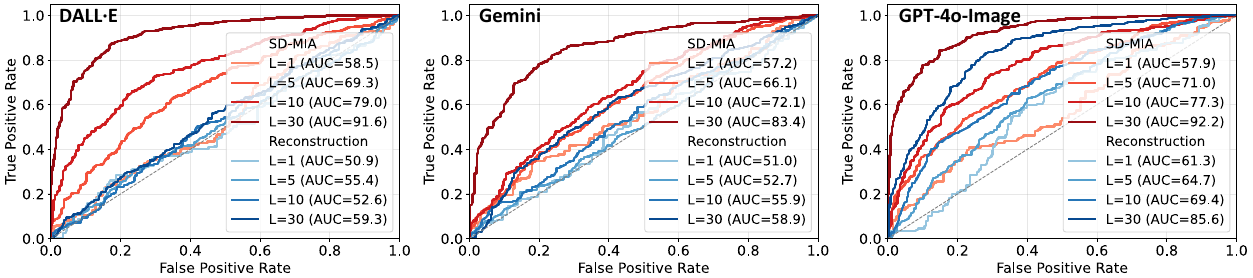}
    \caption{Membership inference attack performance against closed-source image generation models (dall-e-3, gemini-2.0, gpt-4o).}
    \label{fig:API_set}
\end{figure*}


\subsection{Main Results}
We summarize the membership inference performance across various text-to-image diffusion models in Table~\ref{Tab:main}. 
First, SD-MIA consistently achieves the highest performance among all baselines because it directly probes the representation-region collapse associated with memorized pre-training samples. By applying structured cross-modal perturbations and pooling maximum relevance across multiple stochastic generations, SD-MIA exposes stability patterns that are inaccessible to traditional image-space or denoising-based attacks. This enables SD-MIA to outperform even the strongest gray-box baseline, DRC~\cite{fu2025unlocking}, by up to 10\% AUC, demonstrating the advantage of using perturbation-driven cross-modal consistency as a discriminative membership signal.
Second, SD-MIA exhibits strong robustness under varying positive-to-negative sample ratios. For instance, on Stable-diffusion-v1-4 (1:10), SD-MIA achieves 66.12\% AUC, consistently surpassing all baselines. This indicates that maximum cross-modal relevance pooling reliably preserves membership-dependent signal strength even in highly imbalanced scenarios.
Third, the performance advantage of SD-MIA is maintained across multiple models. This cross-model consistency reflects the general nature of representation-region collapse in diffusion pre-training and indicates that the proposed framework can effectively capture memorization signals across diverse diffusion models for pre-training  membership inference.

We also extend our analysis from instance-level to set-level membership inference to assess the effectiveness of SD-MIA in practical scenarios. As illustrated in Figure~\ref{fig:set-level}, we show the AUC results and dataset inference p-values following~\cite{dubinski2025cdi}, for different set sizes $L$ to examine how aggregation over multiple samples influences the overall performance. The set-level setting better reflects realistic use cases, where the objective of the  adversary is not merely to identify individual training samples, but rather to determine whether an entire dataset has contributed to the pre-training of the target model. 
We observe a consistent performance gain as $L$ increases, with the integration of SD-MIA features significantly enhancing dataset-level MIA and outperforming existing baselines. Notably, when the set size reaches $L=30$, SD-MIA achieves an AUC exceeding 95\%, highlighting that membership signals accumulated from multiple perturbed cross-modal interactions are highly consistent and strongly amplify the separability between member and non-member sets. These results demonstrate that SD-MIA not only captures instance-level memorization but also scales reliably to dataset-level auditing.

\subsection{Evaluation on Closed-source Models}
To further assess the generalization capacity of SD-MIA, we evaluate the performance of black-box methods on leading closed-source image generation models, DALL-E, Gemini and GPT-4o (Figure~\ref{fig:API_set}) based on FlickrMIA-25. 
First, despite these constraints, SD-MIA consistently maintains strong performance across all three APIs and surpasses SOTA black-box baselines~\cite{pangblack}. This suggests that the method captures a modality-level behavioral property of modern large-scale generative models rather than relying on architectural artifacts specific to diffusion models. Second, the results show that reliable pre-training membership inference based on SD-MIA remains feasible even in such severely restricted black-box setting, underscoring the practical significance of cross-modal perturbation mechanisms for real-world auditing and compliance scenarios.

\begin{figure*}
    \centering
    \includegraphics[width=0.99\linewidth]{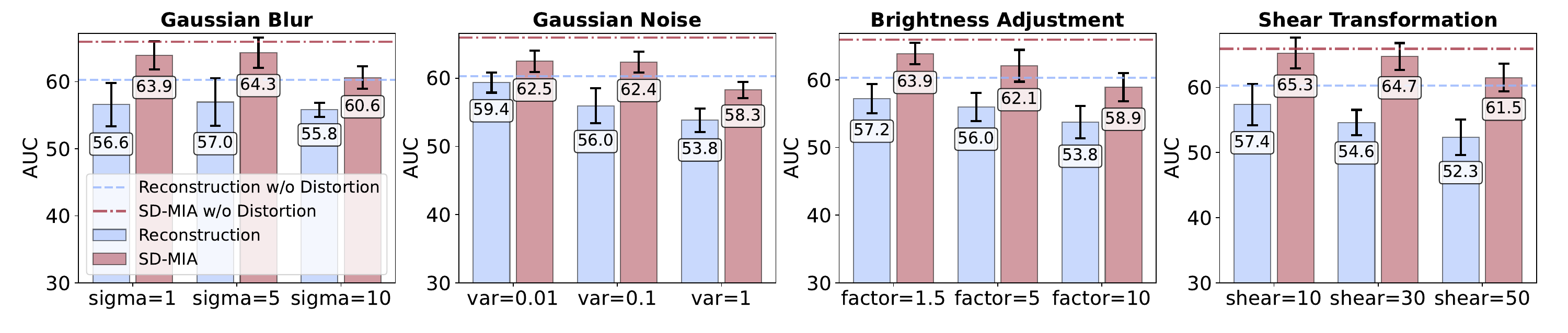}
    \caption{Robustness evaluation under four training-data distortion settings, image blur, Gaussian noise, brightness adjustment, and shear transformation. Within each sub-plot, the distortion intensity increases from left to right.
    }
    \label{fig:robustness}
\end{figure*}

\begin{figure}
    \centering
    \includegraphics[width=.95\linewidth]{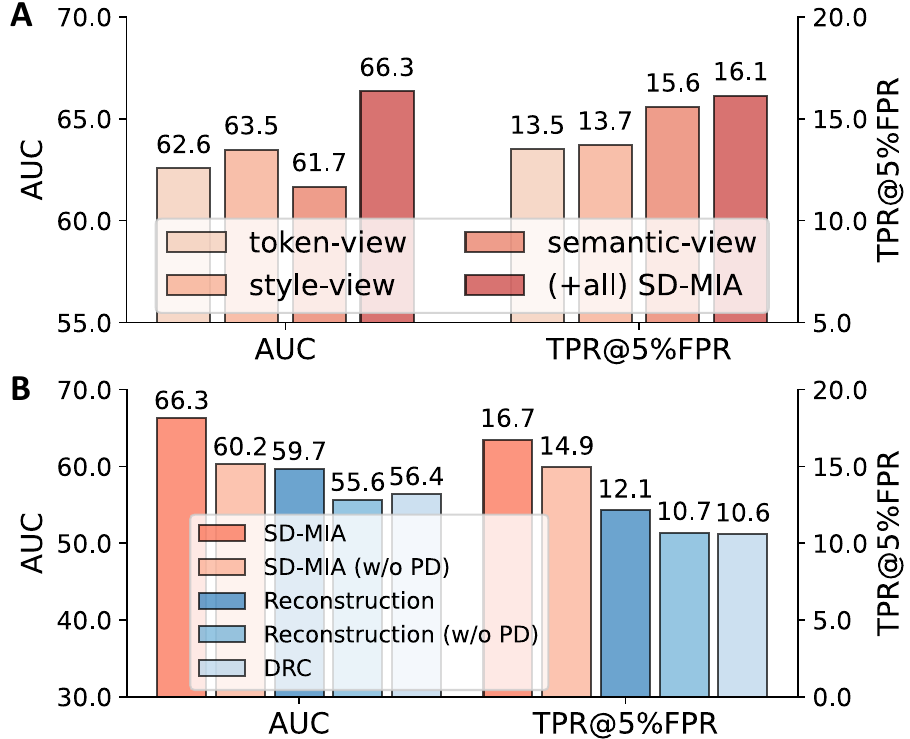}
    \caption{Ablation study of SD-MIA, including: (A) perturbation view of textual input, and (B) paired textual description (PD).}
    \label{fig:ablation}
\end{figure}


\subsection{Robustness Analysis}
To evaluate the robustness of SD-MIA framework, we investigate how controlled variations in the target image affect the performance of MIA methods. Specifically, we apply a series of common geometric and photometric transformations to the suspected original images, including gaussian noise injection, penumbra effects, shear, and brightness adjustment (Figure~\ref{fig:robustness}).
Compared with the baseline Reconstruction method~\cite{pangblack}, SD-MIA demonstrates consistently stronger resilience across all perturbation types. For example, under Gaussian blur perturbation, SD-MIA achieves an AUC of 61.5\%, while the Reconstruction baseline drops to near-random performance.
These findings confirm that SD-MIA maintains reliable membership inference accuracy even under realistic visual distortions, highlighting the inherent stability of its cross-modal perturbation mechanism and its robustness to input disturbance.


%

\subsection{Ablation Study of the SD-MIA Method }
To evaluate the contribution of each perturbation view in the SD-MIA framework, we conduct a detailed ablation study (Figure~\ref{fig:ablation} (A)). 
Specifically, we isolate the three perturbation types and measure the membership inference performance when each type is applied individually. 
This analysis allows us to quantify the relative importance of fine-grained lexical changes, stylistic variations, and semantic modifications in capturing representation-region collapse and discriminative membership signals. The results show that all three perturbation views contribute positively to attack performance. Among them, token-view perturbations, which introduces the smallest embedding shifts, perform well on certain models, demonstrating their sensitivity to subtle memorization patterns. When integrated, the three perturbation types complement one another, enabling SD-MIA to achieve better performance. This confirms that multi-view perturbation is crucial for achieving stable and generalizable membership inference across different models.

We further examine the effect of access to the original paired training description for a suspected image. When the original description is unavailable, we generate a surrogate using a captioning model (e.g., BLIP~\cite{li2023blip}) and apply the same SD-MIA perturbation strategy. The average results, summarized in Figure~\ref{fig:ablation} (B), indicate that while having the paired description provides slightly stronger membership signals, SD-MIA remains highly effective, outperforming the image-only DRC~\cite{fu2025unlocking} baseline. This demonstrates the applicability of SD-MIA in realistic scenarios.
Furthermore, SD-MIA demonstrates a favorable efficiency–utility trade-off, achieving substantial improvements over baseline methods within comparable running times. Due to space limitations, detailed results are provided in the appendix.


\section{Conclusion}
In this work, we introduce SD-MIA, a black-box membership inference framework that exploits cross-modal perturbations to identify pre-training data in diffusion-based image generators.
By employing a multi-view textual perturbation strategy, SD-MIA effectively probes the generation probability curvature associated with a suspected image, which remains largely inaccessible to conventional image-space perturbation methods.
We further propose a maximum cross-modal relevance estimator that quantifies perturbation-induced probability shifts and amplifies the discriminative signals between member and non-member samples.
Comprehensive experiments on two benchmarks and seven image generation models show that SD-MIA consistently outperforms SOTA baselines. Additional analyses highlight the robustness and generalization of our framework, underscoring its practical utility for auditing pre-training data of image generation models.


%

\section*{Acknowledgments}

This work is supported by the National Natural Science Foundation of China under Grant 62502044, 62425203; Beijing Natural Science Foundation under Grant number L253005; CCF-SANGFOR Research Fund under Grant number 20240202; Research Initiation Project for Introduced Talents of BUPT under Grant number 2025KYQD11; and the Beijing Municipal Science \& Technology Commission, the Administrative Commission of Zhongguancun Science Park under Grant number Z251100003625014.
{
    \small
    \bibliographystyle{ieeenat_fullname}
    \bibliography{main_new}
}

\clearpage
\section{Appendix}
\subsection{Preliminary}

Diffusion-based image generation models~\cite{ramesh2022hierarchical,zhang2023adding,saharia2022photorealistic} aim to generate an image $x \in \mathbb{R}^{H \times W \times 3}$ conditioned on a text prompt $c$, by learning the joint distribution $p(x, c)$ or conditional distribution $p(x|c)$ via a denoising diffusion process in either pixel or latent space.  
Formally, the forward diffusion process is defined as:
\begin{equation}
q(z_t|z_{t-1}) = \mathcal{N}\!\left(z_t; \sqrt{1 - \beta_t}z_{t-1}, \beta_t \mathbf{I}\right),
\end{equation}
where $\{\beta_t\}_{t=1}^T$ denotes the noise schedule, and $z_0$ corresponds to the image $x$ or its latent representation $\mathcal{E}(x)$ encoded by a variational autoencoder.  
The reverse process is parameterized by a neural network $\epsilon_\theta(z_t, c, t)$ predicting the noise conditioned on the text embedding $c$:
\begin{equation}
p_\theta(z_{t-1}|z_t, c) = 
\mathcal{N}\!\left(
z_{t-1}; 
\frac{1}{\sqrt{1 - \beta_t}}(z_t - \beta_t \epsilon_\theta(z_t, c, t)), 
\tilde{\beta}_t \mathbf{I}
\right).
\end{equation}
The model is trained on large-scale text–image pairs by minimizing the denoising score matching loss:
\begin{equation}
\mathcal{L}(\theta) =
\mathbb{E}_{z_0, c, t, \epsilon}\!
\left[
\lVert \epsilon - \epsilon_\theta(z_t, c, t) \rVert_2^2
\right],
\end{equation}
where $(z_0, c)$ are sampled from the  training distribution $\mathcal{D}_{\text{train}}$ of text-image pairs.
Through this optimization, the model learns cross-modal correspondences between textual semantics and visual representations. 
Such tightly coupled associations can lead to memorization of specific training pairs, which can subsequently be exploited for MIAs.

\subsection{Baseline Methods}
\label{App:baseline}
Membership inference attacks on image-generation models have recently attracted increasing attention, aiming to determine whether a given sample was involved in model training.
Depending on the adversary's level of access to the target model, existing MIAs can be broadly categorized into white-box, gray-box, and black-box paradigms (Table~\ref{Tab:relatedwork}).
Representative approaches under these settings include the following:

\begin{itemize}
\item Loss~\cite{matsumoto2023membership}: A loss-based attack that exploits the observation that diffusion models yield lower denoising or reconstruction losses for members than for non-members.
\item SecMIA~\cite{duan2023diffusion}: A query-based approach that infers membership by evaluating how well the model's step-wise denoising matches the forward process posterior estimation compared to non-members.
\item PIA~\cite{kongefficient}:  A proximal-initialization-based approach that initializes the diffusion process with optimized noise and traces the denoising trajectory of candidate samples.
\item  CLiD~\cite{zhai2024membership}: A conditional-likelihood-based method that measures the discrepancy between conditional and marginal likelihoods in text-to-image diffusion models, capturing overfitting of conditional distributions as a strong membership signal.

\item DRC~\cite{fu2025unlocking}: A degrade–restore–compare framework that deliberately corrupts salient regions of an input image, restores them using the target diffusion model, and compares the reconstruction quality to infer membership.
\item REDIFFUSE~\cite{li2025towards}: A black-box MIA that uses the diffusion model's image-to-image or variation API. It repeatedly inputs a candidate sample for slight modification, averages the outputs, and compares the reconstruction consistency to the original image to infer membership.
\item Reconstruction~\cite{pangblack}: A black-box reconstruction-based attack leveraging image-to-image or variation APIs to repeatedly regenerate candidate samples and measure reconstruction consistency.
\end{itemize}

\begin{table}[h]
\centering
\caption{Details of MIA methods against diffusion models, where AT, PD, and CM denote the attack type, pre-training data-based evaluation, and use of cross-modal cues for detection, respectively.
}
\label{Tab:relatedwork}
\resizebox{0.46\textwidth}{!}{%
\begin{tabular}{c|ccc}
\toprule
Method                                                    & AT & CM & PD   \\ \midrule
Loss (Matsumoto et al., 2023) ~\cite{matsumoto2023membership}& White        & $\times$                & $\times$        \\ 
SecMI (Duan et al., 2023)~\cite{duan2023diffusion}    & Gray         & $\times$                & $\times$                   \\ 
PIA (Kong et al., 2024)~\cite{kongefficient}          & Gray         & $\times$                 &   $\checkmark$                             \\

CLiD (Zhai et al., 2024)~\cite{zhai2024membership} & Gray         &     $\times$         & $\checkmark$                  
\\
DRC (Fu et al., 2025) ~\cite{fu2025unlocking}& Gray         &   $\times$          &  $\checkmark$                    
\\
REDIFFUSE (Li et al., 2025) ~\cite{li2025towards}   & Black        & $\times$              & $\times$                       \\ 
Reconstruction (Pang and Wang, 2025)~\cite{pangblack} & Black        & $\times$        & $\times$                       \\ \midrule
SD-MIA& Black        & $\checkmark$              &$\checkmark$                  \\ \bottomrule
\end{tabular}
}
\end{table}

\subsection{Experimental Environments}
All experiments were conducted on a high-performance computing server running Ubuntu 22.04.5 LTS. The hardware configuration includes 128 Intel(R) Xeon(R) Gold 6342 CPUs @ 2.80GHz
and 2 NVIDIA A800 80GB PCIe GPUs, providing substantial computational capacity for training and inference. The implementation of all methods was carried out using Python 3.12.9 and PyTorch 2.6.0, which served as the primary framework for model development and experimentation.

\begin{figure}[ht]
    \centering
\includegraphics[width=0.88\linewidth]{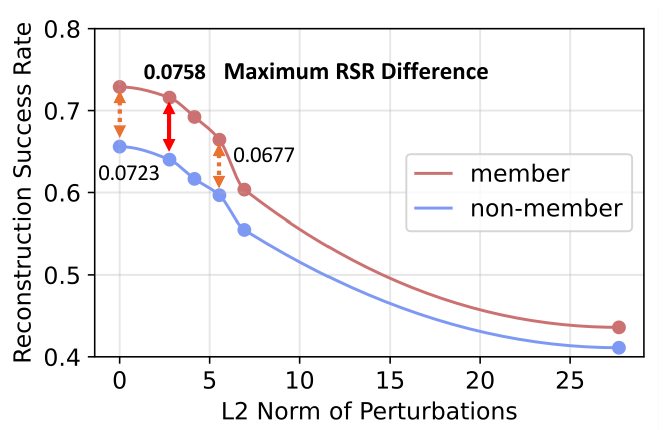}
    \caption{Gray-box evidence of representation collapse.}
    \label{fig.representation_collapse}
\end{figure}

\subsection{Evidence of Representation Collapse}
We provide gray-box evidence of representation collapse and analyze how SD-MIA probes it.
First, we directly perturb embeddings of members and non-members and measure the reconstruction success rate (RSR). Figure~\ref{fig.representation_collapse} shows: under small perturbations ($|\delta|<2.5$), RSR of members remains close to the unperturbed case, while that of non-members drops sharply, indicating that members lie in a locally collapsed region.
Second, under black-box access, SD-MIA perturbs textual prompts to indirectly induce embedding variations. Results (on SD v1-3, v1-4, and v1-5) show that the induced perturbations have an average norm of 0.3858 and a cosine similarity of 0.0066, falling within the validated collapse region above, confirming the effectiveness of the proposed SD-MIA method.

\subsection{Efficiency-Utility Trade-off}
\begin{figure}[th]
       \centering
    \includegraphics[width=0.8\linewidth]{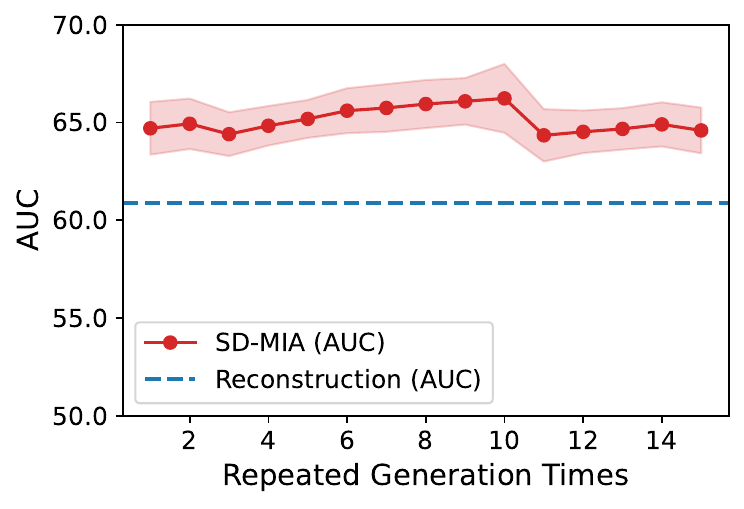}
    \caption{Efficiency-utility trade-off of SD-MIA compared to the Reconstruction baseline, evaluated across different numbers of repeated generations.}
    \label{fig:efficiency} 
\end{figure}

We evaluate the computational efficiency of SD-MIA alongside its membership inference performance to quantify the trade-off between runtime and utility. Figure~\ref{fig:efficiency} presents the membership inference attack performance of SD-MIA and the Reconstruction under varying numbers of repeated generations. The results demonstrate that, with various computational budgets, SD-MIA consistently outperforms the baseline methods, achieving better AUC. This highlights the practicality and efficiency of SD-MIA for large-scale auditing of diffusion models, where maintaining a balance between computational cost and inference accuracy is crucial.

\subsection{Performance on Image Autoregressive Models}
Since SD-MIA is a black-box method that relies only on generated images and avoids diffusion-specific features, it readily transfers to image autoregressive models (IARs), whereas many gray-box methods cannot.
We use Infinity\footnote{\url{https://huggingface.co/FoundationVision/Infinity}} as the target IAR model in our experiments.
Table~\ref{tab:IARs} shows that SD-MIA effectively identifies training data of IAR models and outperforms black-box baselines.

\begin{table}[htbp]
\centering
\caption{Performance on a representative IAR model (Infinity).}
\resizebox{0.49\textwidth}{!}{%
\begin{tabular}{c|cccc}
\toprule
Method & AUC & TPR@1\% &  TPR@5\%  &  TPR@10\% \\
\midrule
Reconstruction & 55.55$_{\pm 2.10}$ & 3.20$_{\pm 0.81}$ & 7.80$_{\pm 2.60}$ & 14.00$_{\pm 4.04}$ \\
SD-MIA & \textbf{58.68}$_{\pm 1.94}$ & \textbf{6.20}$_{\pm 3.42}$ & \textbf{15.20}$_{\pm 2.41}$ & \textbf{22.80}$_{\pm 1.54}$  \\
\bottomrule
\end{tabular}
}
\label{tab:IARs}
\end{table}

\subsection{Experimental Settings for Benchmark Unbiasedness Visualization}
\label{App:Unbiasedness}
To validate the unbiasedness of our benchmarks, we first visualize their embedding-level distribution. Specifically, we encode all samples using CLIP ViT-L/14~\cite{radford2021learning}, then apply Principal Component Analysis (PCA) to reduce the embedding dimensionality to two and plot the resulting distributions, as shown in Figure~\ref{fig:data_distribution}. The visualization indicates that the embeddings of member and non-member samples exhibit well-aligned distributions, suggesting no observable bias.
To further quantify this absence of bias, we train a logistic regression classifier to distinguish between member and non-member embeddings. The classifier uses a regularization strength of 1000 to mitigate overfitting and allows up to 1000 iterations to ensure convergence under high-dimensional, strongly regularized conditions. The resulting classification accuracies are nearly equivalent to random guessing (0.518 for FlickrMIA-25 and 0.533 for LAION-mi), which confirms that our benchmarks are effectively unbiased.

\subsection{Prompts for Perturbation Generation}
\label{App:prompt}
In this section, we present the prompts used to generate multi-view textual perturbations within the SD-MIA framework. These prompts are designed to guide the language model in producing variations at three levels: token-level, style-level, and semantic-level. While introducing these perturbations, the prompts ensure that each part of the original description is preserved, allowing for controlled modifications in both the lexical and stylistic elements of the text.

\begin{promptbox}[Prompt 1: token-view perturbation]
\label{prompt:token-view}
Rewrite the given image caption by {rephrasing} the text while preserving both the original content/subject and the artistic style exactly.
Do not change the main subject or any style modifiers (e.g., 'photorealistic', 'oil painting', 'cartoon style'). Only modify wording, word order, or small descriptive phrasing. \\
Examples: \\
- 'photorealistic, a cat on a chair' → 'photorealistic, a cat sitting on a chair' \\
- 'oil painting of mountains at sunset' → 'oil painting of mountain peaks at sunset' \\
- 'cartoon style, child playing' → 'cartoon style, a child at play' \\
- 'digital art, futuristic cityscape' → 'digital art, a futuristic city skyline' \\
Rules: 1) Preserve the exact subject/content and any style modifiers. Do not introduce new subjects or styles. 2) Only rephrase or slightly rearrange words; avoid adding new objects or changing factual content. 3) Output only the new caption, no quotes or extra text. 4) Ensure the output remains truthful and consistent with the original caption.
\end{promptbox}

\begin{promptbox}[Prompt 2: style-view perturbation]
\label{prompt:style-view}
Rewrite the given image caption so that the content/subject remains exactly the same, but change the artistic style of the image.
Add only 1-2 style modifiers like 'photorealistic', 'cinematic', 'oil painting', 'cartoon style', etc. before, after, or within the caption.\\
Examples:\\
- 'a cat on a chair' → 'photorealistic, a cat on a chair'\\
- 'UK Active logo' → 'UK Active logo, in the style of oil painting'\\
- 'person smiling' → 'a watercolor painting of person smiling'\\
- 'sunset over mountains' → 'cinematic, sunset over mountains'\\
- 'Salad with chestnuts' → 'Salad with chestnuts, digital art'\\
Common style modifiers (choose 1-2 only):\\
- photorealistic, cinematic, highly detailed, 4k\\
- oil painting, watercolor painting, acrylic painting\\
- pencil sketch, ink drawing, charcoal drawing  \\
- cartoon style, anime style, manga\\
- digital art, 3D render, vector art\\
- in the style of [artist/movement]\\
Rules: 1) Keep the exact same content/subject. 2) Add only 1-2 style modifiers (not more). 3) Output only the new caption, no quotes or extra text. 4) Ensure that the output caption conforms to objective facts.
\end{promptbox}

\begin{promptbox}[Prompt 3: semantic-view perturbation]
\label{prompt:semantic-view}
Rewrite the given image caption so that the content/subject is changed, but keep the same artistic STYLE.\\
Keep the same style modifiers (if any) but change the main subject/content.\\
Examples:\\
- 'photorealistic, a cat on a chair' → 'photorealistic, a dog on a sofa'\\
- 'UK Active logo' → 'Nike logo' (both simple descriptions without style modifiers)\\
- 'oil painting of mountains' → 'oil painting of an ocean'\\
- 'sunset over mountains, digital art' → 'sunrise over cityscape, digital art'\\
- 'person smiling' → 'person running' (both simple, no style modifiers)\\
Rules: 1) Change the subject/content to something different. 2) Keep the same style modifiers if present in the original. 3) If no style modifiers in original, keep the same simple format. 4) Output only the new caption, no quotes or extra text. 5) Ensure that the output caption conforms to objective facts.
\end{promptbox}





\end{document}